\documentclass[11pt,a4paper]{article}
\usepackage[hyperref]{acl2021}
\usepackage[pdftex]{}
\pdfoutput=1
\usepackage{times}
\usepackage{latexsym}

\usepackage{booktabs}
\usepackage{graphicx}
\usepackage{enumitem}

\usepackage{microtype}

\aclfinalcopy 

\title{Predicting Different Types of Subtle Toxicity in Unhealthy Online Conversations}

\usepackage{tcolorbox}

\newcommand{\takeaway}[1]{\noindent
    \begin{tcolorbox}[]
    {#1}
    \end{tcolorbox}
}

\author{
  Shlok Gilda \\
  University of Florida\\
  \texttt{shlokgilda@ufl.edu} \\\And
  Mirela Silva\\
  University of Florida\\
  \texttt{msilva1@ufl.edu} \\\AND
  Luiz Giovanini\\
  University of Florida\\
  \texttt{lfrancogiovanini@ufl.edu} 
  \\\And
  Daniela Oliveira\\
  University of Florida \\ 
  \texttt{daniela@ece.ufl.edu} \\
  }

\date{\today}

\begin{document}
\maketitle

\begin{abstract}
    
This paper investigates the use of machine learning models for the classification of unhealthy online conversations containing one or more forms of subtler abuse, such as hostility, sarcasm, and generalization. We leveraged a public dataset of 44K online comments containing healthy and unhealthy comments labeled with seven forms of subtle toxicity. We were able to distinguish between these comments with a top micro F1-score, macro F1-score, and ROC-AUC of 88.76\%, 67.98\%, and 0.71, respectively. Hostile comments were easier to detect than other types of unhealthy comments. We also conducted a sentiment analysis which revealed that most types of unhealthy comments were associated with a slight negative sentiment, with hostile comments being the most negative ones.

\end{abstract}

\section{Introduction}\label{sec:intro}

Healthy online conversations occur when posts or comments are made in good faith, are not blatantly abusive or hostile, typically focus on substance and ideas, and generally invite engagement~\cite{Price_Gifford-Moore_Flemming_Musker_Roichman_Sylvain_Thain_Dixon_Sorensen_2020}. 
Conversely, toxic comments are a harmful type of conversation widely found online which are insulting and violent in nature~\cite{Price_Gifford-Moore_Flemming_Musker_Roichman_Sylvain_Thain_Dixon_Sorensen_2020}, such as ``SHUT UP, YOU FAT POOP, OR I WILL KICK YOUR A**!!!". 
This kind of toxic conversation has been the primary focus of several previous studies~\cite{Georgakopoulos_Tasoulis_Vrahatis_Plagianakos_2018, Srivastava_Khurana_Tewari}. However, many comments that deter people from engaging in online conversations are not necessarily outright abusive, but contain subtle forms of abuse (e.g., ``Because it drives you crazy, old lady. Toodooloo..."). 
These comments are written in a way to engage people, but to also hurt, antagonize, or humiliate others and are thus referred to as \emph{unhealthy conversations}~\cite{Price_Gifford-Moore_Flemming_Musker_Roichman_Sylvain_Thain_Dixon_Sorensen_2020}. 
In other words, unhealthy comments are less negative, intense, and hostile than toxic ones. Detecting unhealthy conversations is more challenging and less explored in the literature than its toxic conversations counterpart~\cite{Price_Gifford-Moore_Flemming_Musker_Roichman_Sylvain_Thain_Dixon_Sorensen_2020}.

Many early signs of conversational failure are due to the subtlety in comments which deter people from engagement and create downward spirals in interactions. 
Behaviors such as condescension (e.g., ``Utter drivel and undeserving of further response."), ``benevolent'' stereotyping (e.g., ``Women have motherly nurturing instincts."), and microaggressions (e.g., racially-based such as ``Why don't you have an accent?") are frequently experienced by members of minority social groups~\cite{sue2007racial, glick2001ambivalent}. 
\citet{Nadal_Griffin_Wong_Hamit_Rasmus_2014} indicated that such subtle abuse can be as emotionally harmful as outright toxic abuse to some individuals. 
Microaggressions (even unintended slights or social cues) have been linked to disagreements in intergroup relationships.
\citet{Jurgens_Chandrasekharan_Hemphill_2019} also signify the importance of tackling more subtle and serious forms of online abuse by developing proactive technologies (e.g., intervention by bystanders [\citealp{Markey_2000}], rephrasing parts of a message to adjust the level of politeness [\citealp{politeness}]) to counter abuse before it can cause harm.

In this paper, we sought to answer two research questions in the context of unhealthy conversations:
\begin{itemize}
    \item {\textbf{RQ1:}} What is the general sentiment associated with unhealthy conversations compared to healthy conversations? 
    \item {\textbf{RQ2:}} Can we differentiate unhealthy and healthy conversations? If so, which type of unhealthy conversation is the most detectable?
\end{itemize}

Towards this end, we analyzed a dataset comprised of 44K labeled comments of unhealthy conversations from \citet{Price_Gifford-Moore_Flemming_Musker_Roichman_Sylvain_Thain_Dixon_Sorensen_2020}. 
We submitted this dataset first to sentiment analysis aimed at detecting the polarity (positive/negative/neutral) of the comments (\textbf{RQ1}), and then to a comprehensive machine learning analysis focused on distinguishing comments as healthy vs. unhealthy. (\textbf{RQ2}). Our experimental results revealed that: \textbf{(i) although none of the types of comments had extremely polarizing sentiments, most forms of unhealthy online conversations were associated with a slight negative sentiment;} and \textbf{(ii) hostile comments were the most detectable form of unhealthy conversation}. 
Findings from this work have the potential to inform and advance future research and development of online moderation tools, which pave the way for safer online environments.


This paper is organized as follows. 
Section~\ref{sec:relatedwork} summarizes related work. 
Section~\ref{sec:dataset} describes the dataset we leveraged for our experiments, as well as our preprocessing procedures. 
Section~\ref{sec:methodology} details our study's methodology. 
Section~\ref{sec:discussion} analyzes our study's results, and discusses our results by answering our proposed research questions.
Section~\ref{sec:limitations} analyzes our study's limitations and proposes directions for future work. 
Section~\ref{sec:conclusion} concludes the paper.

\section{Related Work}\label{sec:relatedwork}


Sentiment classification of social media posts with regards to toxicity has been researched extensively over the past years~\cite{Chen_Qian_2019,Saeed_Shahzad_Kamiran_2018,Saif_Medvedev_Medvedev_Atanasova_2018,Srivastava_Khurana_Tewari}. The primary focus of most of the related work has been on algorithmic moderation of toxic comments, which are derogatory and threatening. The importance of community norms in detection and classification of these subtler forms of abuse have been noted elsewhere~\cite{Blackwell_Dimond_Schoenebeck_Lampe_2017,Guberman_Hemphill_2017,Liu_Guberman_Hemphill_Culotta_2018,Salminen_Almerekhi_Milenkovic_Jung_An_Kwak_Jansen_2019}, but have not received the same attention in the NLP community.

Although recognized in the larger NLP abuse typology~\cite{Waseem_Davidson_Warmsley_Weber_2017}, there have been only a few attempts at solving the problems associated with subtle abuse detection, such as a study on classification of ambivalent sexism using Twitter data~\cite{Jha_Mamidi_2017}. 
There is a need for development of new methods to identify the implicit signals of conversational cues. Detecting subtler forms of toxicity requires idiosyncratic knowledge, familiarity with the conversation context, or familiarity with the cultural tropes~\cite{van_Aken_Risch_Krestel_Loser_2018,Parekh_Patel_2017}. 
It also requires reasoning about the implications of the propositions. 
\citet{Dinakar_Jones_Havasi_Lieberman_Picard_2012} extract implicit assumptions in statements and use common sense reasoning to identify social norm violations that would be considered an insult.

Identification of subtle indicators of unhealthy conversations in online comments is a challenging task due to three main reasons~\cite{Price_Gifford-Moore_Flemming_Musker_Roichman_Sylvain_Thain_Dixon_Sorensen_2020}: (i) comments are less extreme and thus have lesser explicit vocabulary; (ii) a remark may be perceived differently based on context or expectations of the reader; and (iii) greater risk of false positives or false negatives. 
Cultural diversity also plays an important role on how a comment/remark may be perceived differently~\cite{qiufen2014understanding}, thus making identification of subtle toxicity online more challenging.

From a machine learning perspective, abusive comments classification research initially began with the application of combining TF-IDF (Term Frequency–Inverse Document Frequency) with sentiment and contextual features by~\citet{Yin_Xue_Hong_Davison_Edwards_Edwards_2009}. They compared the performance of this model with a simple TF-IDF model and reported a 6\% increase in the F1-score of the classifier on chat-style datasets. Since then, there have been many advances in the field of toxicity classification. \citet{Davidson_Warmsley_Macy_Weber_2017} detected hate speech on Twitter using a three-way classifier: hate speech, offensive but not hate speech, and none. They implemented Logistic Regression and SVM classifiers using several different features including TF-IDF n-grams, Part-of-Speech (POS) $n$-grams, Flesch Reading Ease scores, and sentiment scores using an existing sentiment lexicon. Using a similar set of features, \citet{Safi_Samghabadi_Maharjan_Sprague_Diaz-Sprague_Solorio_2017} applied a linear SVM to detect ``invective" posts on Ask.fm, a social networking site for asking questions. They also utilized additional features such as Linguistic Inquiry and Word Count (LIWC; \citealp{pennebaker01}), word2vec~\cite{Mikolov_Chen_Corrado_Dean_2013}, paragraph2vec~\cite{Le_Mikolov_2014}, and topic modeling. The authors reported an F1-score of 59\% and AUC-ROC (area under the ROC curve) of 0.785 using a specific subset of features. 
\citet{Yu_Wang_Lai_Zhang_2017} proposed a word vector refinement model that could be applied to pre-trained word vectors (e.g., Word2Vec [\citealp{Mikolov_Chen_Corrado_Dean_2013}] and Glove [\citealp{Pennington_Socher_Manning_2014}]) to improve the efficiency of sentiment analysis. 
\citet{Chu2017CommentAC} compared the performance of various deep learning approaches to toxicity classification, using both word and character embeddings. They analyzed the performance of neural networks with LSTM (Long Short-Term Memory) and word embeddings, a CNN (Convolution Neural Network) with word embeddings, and another CNN with character embeddings; the latter achieved top accuracy of 93\%. 

This paper aims to tackle the issue of recognizing unhealthy online conversations. Existing research usually focuses on classifying the most hateful/vile comments; we, on the other hand, aimed to identify the subtle toxicity indicators in online conversations.

\section{Data Preparation}\label{sec:dataset}
In this section, we describe the dataset of 44K labeled comments of unhealthy conversations from \citet{Price_Gifford-Moore_Flemming_Musker_Roichman_Sylvain_Thain_Dixon_Sorensen_2020}, and detail the preprocessing steps taken to prepare the dataset for our machine learning analyses.

\subsection{Dataset Description}
The dataset used in this study was made publicly available\footnote{\url{https://github.com/conversationai/unhealthy-conversations}} by \citet{Price_Gifford-Moore_Flemming_Musker_Roichman_Sylvain_Thain_Dixon_Sorensen_2020} in October 2020. It contains 44,355 unique comments of 250 characters or less from the Globe and Mail\footnote{\url{https://www.theglobeandmail.com/}} opinion articles sampled from the Simon Fraser University Opinion and Comments Corpus dataset by \citet{Kolhatkar_Wu_Cavasso_Francis_Shukla_Taboada_2020}. 
Each comment was coded by at least three annotators with at least one of the following class labels: \textit{antagonize, condescending, dismissive, generalization, generalization unfair, healthy, hostile,} and \textit{sarcastic}. 
A maximum of five annotators were used per comment until sufficient consensus---measured by a confidence score $\geq$ 75\%---was reached. Each annotator was asked to identify for each comment whether it was healthy and if any of the types of unhealthy discourse (i.e., \textit{condescending, generalization}, etc.) were present or not. The comments were presented in isolation to annotators, without the surrounding context of the news article and other comments (i.e., the annotators had no additional context about where the comment was posted or about the engagement of the comment with the users), thus possibly reducing bias. Since the comments were sourced from the SFU Opinion and Comments Corpus dataset, the prevalence of each attribute is unavoidably low.

The following are a few examples of every class label:
\begin{enumerate}[leftmargin=*, nosep]
\item \emph{Antagonize:} ``An idiot criticizing another idiot, seems about right.''
\item \emph{Condescending:} ``So, everyone who disagrees with her - and YOU - are fascists? You're just another pouting SJW\footnote{Social Justice Warrior; a pejorative term typically aimed at someone who espouses socially liberal movements.}, angry that you didn't get a `Participant' trophy...''
\item \emph{Dismissive:} ``You are certifiably a nut job with that comment.''
\item \emph{Generalization:} ``This is all on the Greeks. Period. They are the ones who have been overspending for decades and lying about it.''
\item \emph{Generalization Unfair:} ``Progressives ALWAYS go over the top. They cannot help it. It's part of their MO.''
\item \emph{Healthy:} ``We are seeing fascist tendencies and behaviours in our government, but its [sic] so hard to believe that it is happening to us, to Canada, that we will just continue to be in denial.''
\item \emph{Hostile:} ``Crazy religious f**kers are the cause of overpopulating the planet with inbred, uneducated people with poor prospects of earning a satisfying life. An endless pool of desperate terrorists in waiting.''
\item \emph{Sarcastic:} ``You should write another comment, [user] - there are a few right-wing buzzwords you didn't use yet. Not many, but some.''
\end{enumerate}

\smallskip
\citet{Price_Gifford-Moore_Flemming_Musker_Roichman_Sylvain_Thain_Dixon_Sorensen_2020} used Krippendorff’s $\alpha$ as the inter-rater reliability metric for the crowd-sourced annotators (note, however, that they did not report the $\alpha$ for two labels: \textit{healthy} and \textit{generalization unfair}). 
Opposite to most inter-rater reliability metrics which address agreement, Krippendorff’s $\alpha$ measures the \emph{disagreement} among coders, ranging from 0 (perfect disagreement) to 1 (perfect agreement). Table~\ref{tab:dataset_preprocessed_krippendorff} lists the Krippendorff’s $\alpha$ for each attribute (reproduced from \citealp{Price_Gifford-Moore_Flemming_Musker_Roichman_Sylvain_Thain_Dixon_Sorensen_2020}), along with the final distribution of our preprocessed dataset (detailed below).

    

\begin{table}[ht]
\caption{Distribution of comments per category along with the respective Krippendorff's $\alpha$ reported by \citet{Price_Gifford-Moore_Flemming_Musker_Roichman_Sylvain_Thain_Dixon_Sorensen_2020}. Note that some comments (10.6\%) were attributed to multiple labels.}
\label{tab:dataset_preprocessed_krippendorff}
\begin{tabular}{|c|c|c|}
\hline
\textbf{Class label} & \textbf{Count} & \textbf{K-alpha} \\ \hline
Antagonize & 2,066 & 0.39 \\ \hline
Condescending & 2,434 & 0.36 \\ \hline
Dismissive & 1,364 & 0.31 \\ \hline
Generalization & 944 & 0.35 \\ \hline
Generalization Unfair & 890 & Not reported \\ \hline
Healthy & 41,040 & Not reported \\ \hline
Hostile & 1,130 & 0.36 \\ \hline
Sarcastic & 1,897 & 0.34 \\ \hline
\end{tabular}
\end{table}

\subsection{Preprocessing}
First, we removed one comment from the dataset that was empty and 1,106 other comments which were not assigned to any class label. This decreased the total number of comments to 43,248. All comments were then preprocessed for the feature extraction step. We converted all characters to lower-case, then removed HTML tags, non-alphabetic characters, newline, tab characters, and punctuation from the comments. We also converted accented characters to their standardized representations to avoid our classifier ambiguating words such as ``latte'' and ``latté.'' 
We then expanded contractions (e.g., ``don't'' is replaced with ``do not''). The final step was lemmatization, which consists of reducing words to their root forms (e.g., ``playing'' becomes ``play''). After preprocessing the data, three comments were deleted because they contained just numbers or special characters. Thus, the total number of comments was \textbf{43,245} with an average length of 19.8 words. The final distribution of the preprocessed dataset is shown in Table~\ref{tab:dataset_preprocessed_krippendorff}. Most of the comments were assigned to a single class label ($N=38,661$, 89.4\%), while 10.6\% of the comments ($N=4,584$) were associated with two or more labels.

\section{Methodology and Analysis}\label{sec:methodology}
This section describes our feature engineering process followed by a detailed description of our machine learning analysis targeting the recognition of unhealthy comments through a diverse set of classification models, including traditional and deep architectures. We also describe the process of sentiment analysis of the comments.

~\subsection{Feature Engineering}
We vectorized the comments using the Term Frequency-Inverse Document Frequency (TF-IDF) statistic, which is commonly used to evaluate how important a word is to a text in relation to a collection of texts~\cite{ir:2008}. This measure considers not only the frequency of words or character $n$-grams in the text but also the relevancy of those tokens across the dataset as a whole. We also included other features that we considered useful in recognizing unhealthy comments, including length of comments, percentage of characters which are capitalized in each comment, and percentage of punctuation characters in each comment.

~\subsection{Machine Learning Analysis}

In our machine learning experiments of multi-label classification, we considered the following well-known models:

\medskip \noindent 
\textbf{Logistic regression}. We used the Logistic Regression model with TF-IDF vectorized comment texts using only words for tokens (limited to 10K features). 

\medskip \noindent
\textbf{Support Vector Machine (SVM)}. 
Essentially, SVM focuses on a small subset of examples that are critical to differentiating between class members and non-class members, throwing out the remaining examples~\cite{Vapnik1998}. This is a crucial property when analyzing large data sets containing many ambiguous patterns. 
We used linear kernel since it is robust to overfitting.

\medskip \noindent 
\textbf{LightGBM}. LightGBM is a tree-based ensemble model that is trained with gradient boosting~\cite{Ke_Meng_Finley_Wang_Chen_Ma_Ye_Liu_2017, barbier_Dia_Macris_Krzakala_Lesieur_Zdeborova_2016,zhang2017gpuacceleration}. The unique attribute of LightGBM versus other boosted tree algorithms is that it grows leaf-wise rather than level-wise, meaning that it prioritizes width over depth. There is an important distinction between boosted tree models and random forest models. While forest models like Scikit-Learn's~\cite{scikit-learn} \emph{RandomForest} use an ensemble of fully developed decision trees, boosted tree algorithms use an ensemble of weak learners that may be trained faster and can possibly generalize better on a dataset like this one where there are a very large number of features but only a select few might have an influence on any given comment. Given the huge disparity between healthy and unhealthy comments, we used a tree-based model\footnote{\url{https://github.com/microsoft/LightGBM}} with the top 50 words and the engineered features.

\begin{figure}[ht]
    \centering
    \hspace*{0mm}                                      
    \includegraphics[width=1.1\columnwidth,keepaspectratio]
    {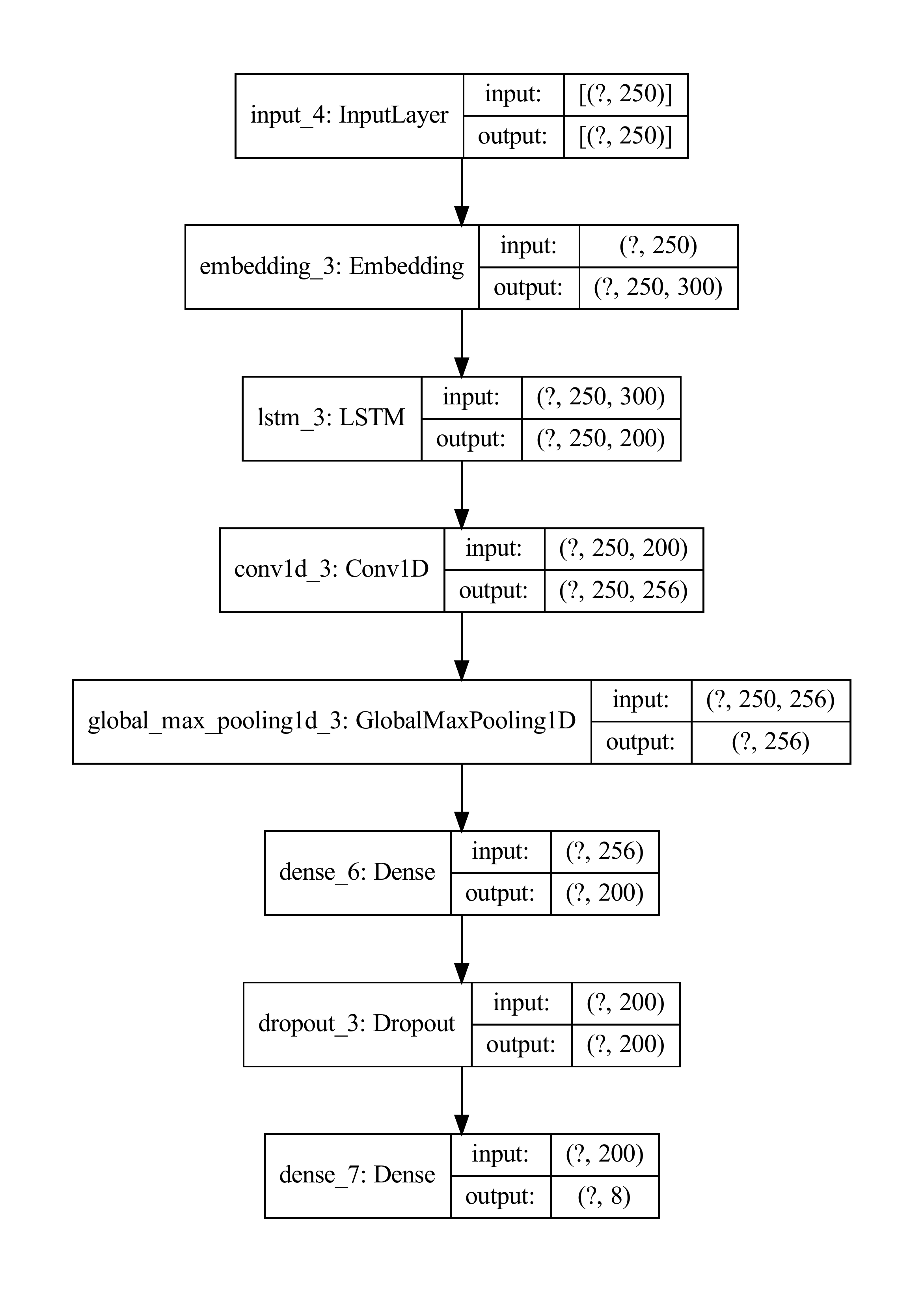}
    \caption{CNN-LSTM network architecture.}
    \label{fig:cnn_lstm_network}
\end{figure}

\medskip \noindent 
\textbf{Convolutional Neural Network Long Short Term Memory (CNN-LSTM) with pre-trained word embeddings}.
We used Global Vectors for Word Representation (GloVe; \citealp{pennington2014glove}) to create an index of words mapped to known embeddings by parsing the data dump of pre-trained embeddings. Since the comments had variable length (range: $[3,250]$ characters), we fixed the comment length at 250 characters, adding zeros at the end of the sequence vector for shorter sentences. Next, the embedding layer was used to load the pre-trained word embedding model. The LSTM layer effectively preserved the characteristics of historical information in long texts, and the following CNN layer extracted the local features from the text. After the CNN, we made use of $global max pooling$ function to reduce the dimensionality of the feature maps output by the preceding convolutional layer. The pooling layer was followed by a dense layer (regular deeply connected neural network); there was also a dropout layer placed between the two dense layers to help regularize the learning and reduce overfitting of the dataset. We used Tensorflow~\cite{tensorflow2015-whitepaper} to implement our CNN-LSTM deep model, whose architecture is illustrated in Fig.~\ref{fig:cnn_lstm_network}. We used binary cross entropy as loss function because it handles each class as an independent vector (instead of as an 8-dimensional vector). The input to the model was a random number of samples (represented as ``$?$'' in Fig.~\ref{fig:cnn_lstm_network}), all having a fixed length of 250 characters.

\medskip
For Logistic Regression, SVM, and LightGBM, we used the implementations available in Scikit-Learn~\cite{scikit-learn} with slight modifications to the default parameters. These models were evaluated using $k$-fold cross-validation for $k = 10$. Importantly, in multi-label classification tasks (our case), a given comment may be associated with more than one label, so we did not use  traditional stratified $k$-fold sampling which preserves the distribution of all classes. Instead, we opted to use iterative stratification~\cite{sechidis2011stratification,pmlr-v74-szymanski17a}, via the \emph{IterativeStratification} method from Scikit-Mulitlearn~\cite{2017arXiv170201460S}, which gives a well-balanced distribution of evidence of label relations up to a given order. 
Owing to the high computational costs of training a deep neural network, the CNN-LSTM neural network was evaluated with 5-fold cross-validation instead of 10 folds.

The evaluation metrics used in our experiments were \textit{micro and macro F1-scores}, and \textit{AUC-ROC} (i.e., area under the ROC curve). F1-score is defined as the harmonic mean of precision, a measure of \emph{exactness}, and recall, a measure of \emph{completeness}. It is well-suited to handle imbalanced datasets (as in our case; \citealp{han2012mining}). A macro-average F1-score will compute the F1-score independently for each class and then take the average (hence treating all classes equally), whereas a micro-average F1-score will aggregate the contributions of all classes to compute the average F1-score~\cite{Zheng_2015}. AUC-ROC curve is a performance measurement for the classification problems at various threshold settings. ROC is a probability curve and AUC represents the degree or measure of separability, indicating how much the model is capable of distinguishing between classes (wherein higher values indicate better discernment; \citealp{Zheng_2015}). 

We ran multiple experiments with different values of hyper-parameters for every model. For the logistic regression model, we tested different values of $n$-grams (words and characters), and max features (words and characters). We achieved the best results with 10,000 word max-features and word $n$-grams range between $[1,3]$. For the SVM model, we chose a linear kernel and tested distinct values of the regularization parameter ($C$); we observed that a value of $C = 0.7$ gave us better results than the default value of $C = 1.0$. During our experiments with LightGBM, we modified some baseline parameters to better suit the problem at hand: $num\_{leaves} = 128$, $n\_{estimators} = 700$ and $max\_{depth} = 32$.

~\subsection{Sentiment Analysis}
We used NLTK VADER (Valence Aware Dictionary for sEntiment Reasoning; \citealp{inproceedings_vader}) to analyze the polarity of comments. VADER is a lexicon and rule-based sentiment analysis tool that is specifically attuned to sentiments expressed in social media. We used the \emph{compound} value from the result for analyzing the polarity of the sentiments. For every input text, VADER normalizes the overall sentiment score to fall within $-1$ (very negative) and $+1$ (very positive), where scores between $(-0.05,0.05)$ are labeled as neutral polarity.

\section{Results \& Discussion}\label{sec:discussion}

In this paper, we analyzed the granularities of subtle toxic online comments. This section presents our experimental results in detecting the general sentiments associated with healthy and unhealthy online comments (\textbf{RQ1}) as well as recognizing such comments via machine learning classifiers (\textbf{RQ2}). Lastly, we summarize our main takeaways.

\subsection{Results}
Table~\ref{tab:sentiment_analysis} shows the mean values of the sentiment polarity for each category of comment. All types of unhealthy comments except \emph{sarcastic} and \emph{condescending} resulted in slight negative scores. The most negative result was observed for the class \textit{hostile}, while the most positive result was obtained from the class \textit{sarcastic}. The classes \textit{condescending} and \textit{healthy} produced the most neutral sentiment scores.



\begin{table}[ht]
\centering
\caption{Sentiment Analysis using VADER's compound score; values range from $-1$ (extremely negative) to $+1$ (extremely positive).}
\label{tab:sentiment_analysis}
\begin{tabular}{|c|c|}
\hline
\textbf{Class label} & \textbf{Mean score} \\ \hline
Antagonize & $-0.104498$ \\ \hline
Condescending & $-0.035128$ \\ \hline
Dismissive & $-0.081356$ \\ \hline
Generalization & $-0.085851$ \\ \hline
Generalization Unfair & $-0.091320$ \\ \hline
Healthy & ~ ~$0.035745$ \\ \hline
Hostile & $-0.175975$ \\ \hline
Sarcastic & ~ ~$0.101285$ \\ \hline
\end{tabular}
\end{table}



\begin{table*}[t]
\centering
\caption{Classification results.}
\label{tab:machine_learning_results}
\begin{tabular}{|c|c|c|c|}
\hline
\textbf{Model} & \multicolumn{1}{l|}{\textbf{Average Micro F1}} & \multicolumn{1}{l|}{\textbf{Average Macro F1}} & \textbf{AUC-ROC} \\ \hline
Logistic Regression & 57.54\% & 48.31\% & 0.51 \\ \hline
SVM & 69.15\% & 61.29\% & 0.62 \\ \hline
LightGBM & 64.83\% & 52.11\% & 0.57 \\ \hline
CNN LSTM Network & \textbf{88.76\%} & \textbf{67.98\%} & \textbf{0.71} \\ \hline
\end{tabular}
\end{table*}

\begin{table}[t]
\caption{AUC-ROC results from the CNN-LSTM model.}
\centering
\label{tab:roc_auc_curve}
\begin{tabular}{|c|c|}
\hline
\textbf{Class label} & \textbf{ROC-AUC} \\ \hline
Antagonize & 0.7362 \\ \hline
Condescending & 0.6702 \\ \hline
Dismissive & 0.6311 \\ \hline
Generalization & 0.6633 \\ \hline
Generalization Unfair & 0.6590 \\ \hline
Healthy & 0.9524 \\ \hline
Hostile & 0.8141 \\ \hline
Sarcastic & 0.5707 \\ \hline
{\textbf{Mean}} & {\textbf{0.7121}} \\ \hline
\end{tabular}
\end{table}

Table~\ref{tab:machine_learning_results} exhibits the average micro F1-score, macro F1-score, and ROC-AUC obtained with all tested classifiers. As can be observed, the best classification results were achieved with the CNN-LSTM model, followed by SVM and LightGBM. Table~\ref{tab:roc_auc_curve} presents the ROC-AUC values resulted from CNN-LSTM (our best-performing classifier) for all analyzed types of comments. The best result was observed for the class \textit{healthy} ($AUC=0.9524$), followed by the classes \textit{hostile} ($AUC=0.8141$) and \textit{antagonize} ($AUC=0.7362$). The class \textit{sarcasm} yielded the poorest result ($AUC=0.5707$).

\subsection{Take-Aways}

\subsubsection*{RQ1: What is the general sentiment associated with unhealthy conversations compared to healthy conversations?}

Our sentiment analysis revealed that none of the types of comments (i.e., classes) have extremely polarizing sentiment values (Table~\ref{tab:sentiment_analysis}). However, all the classes except \textit{healthy}, \textit{sarcastic}, and \emph{condescending} have an overall slightly negative sentiment associated with them, as expected for unhealthy types of conversations. Hostile comments were the most negative, likely due to the use of blatantly vulgar and vile language.
This could possibly indicate that hostile comments were less subtle in their hateful content, and thus easier for sentiment tools and machine learning algorithms to detect.

Antagonizing, dismissive, and generalizing comments all had similar negative sentiment scores in the range of $[-0.1045, -0.0814]$.
Though it is likely clear to most readers that such comments (examples given in Sec.~\ref{sec:dataset}) demonstrate negative sentiment, both NLTK's VADER and CNN-LSTM 
demonstrated poor results for these classes, which highlights how sentiment tools and machine learning models may struggle to detect subtle forms of unhealthy language.

Interestingly, condescending comments scored neutral sentiment ($-0.03513$). Detecting patronizing and condescending language is still an open research problem because, amongst many reasons, condescension is often shrouded under ``flowery words'' and can itself fall into seven different categories~\cite{perezalmendros2020dont}. Meanwhile, sarcastic comments were notably labeled as positive sentiment ($0.1013$), even higher than healthy comments ($0.0357$); this may have been because sarcasm tends to be an ironic remark, often veiled in a potentially distracting positive tone.

\medskip
\takeaway{Unhealthy comments were mostly related with a low polarizing negative sentiment, with \emph{hostile} comments labeled as the most negative and \emph{sarcastic} as the most positive. Healthy comments were associated with neutral sentiment.}

\subsubsection*{RQ2: Can we differentiate unhealthy and healthy conversations? If so, which type of unhealthy conversation is the most detectable one?}

From Table~\ref{tab:machine_learning_results}, we can see that it was possible to differentiate between unhealthy comments with a maximum micro F1-score, macro F1-score, and ROC-AUC of 88.76\%, 67.98\%, and $0.7121$, respectively, using the CNN-LSTM model. The remaining models tested (Logistic Regression, SVM, and LightGBM) reached poorer performances. All tested models reported a lower average macro F1-score compared to the micro F1-score results, which is intuitive given the highly imbalanced dataset---note that macro F1-score gives the same importance to each class, i.e., this value is low for models that perform poorly on rare classes, which was the case for CNN-LSTM when analyzing unhealthy conversation classes. Therefore, it is possible to speculate that the other models also performed better when recognizing healthy comments compared to unhealthy ones.

Additionally, Table~\ref{tab:roc_auc_curve} shows that healthy comments can be differentiated from the remaining classes relatively well with generally high predictive accuracy ($AUC_{healthy}=0.9524$), likely due to the high number of healthy samples in the dataset. In contrast, despite only 1,130  samples associated with the \textit{hostile} class, $AUC_{hostile}=0.8141$, which was the second-highest AUC achieved by our top performer classifier. Most of the hostile comments have explicit language, which might make it easier for the classifier to properly recognize this type of conversation. 
The AUC for antagonizing comments was the third-highest achieved, at $0.7362$, potentially because $n_{antagonize}=2,066$ was the second largest sample size out of the remaining unhealthy conversations categories; \textit{antagonize} also achieved the second most negative mean sentiment score ($-0.1045$), further indicating that there may be characteristics of this class that facilitate detection from machine learning algorithms.

Detection of \textit{sarcasm} was the most difficult for the CNN-LSTM model ($AUC_{sarcasm} = 0.5707$). There have been numerous studies which have faced similar issues with detection of sarcasm given the difficulty of understanding the nuances and context surrounding sarcastic texts~\cite{poria2017deeper, Zhang2016TweetSD}.  The AUC scores for condescending, dismissive, and generalizing comments were only slightly better (range: $[0.6590, 0.6702]$), again highlighting the difficulty in detecting such nuanced and subtle language.

\medskip
\takeaway{Healthy online comments were accurately distinguished from unhealthy ones. Hostile comments were easier to detect than other forms of unhealthy conversations.}

\section{Limitations \& Future Work}\label{sec:limitations}

This section discusses the limitations of our study, and suggests possible future work.

We leveraged the \citet{Price_Gifford-Moore_Flemming_Musker_Roichman_Sylvain_Thain_Dixon_Sorensen_2020} dataset, which sampled comments from the opinions section of the Canadian Globe and Mail news website. Though our results are promising (with top average micro and macro F1-scores of 88.76\% and 67.98\%, respectively), this dataset is nonetheless contained in a highly specific context (data was collected from a single Canadian newspaper website), which likely decreases the generalizability of our results. The dataset was also notably imbalanced, where most of the labels were attributed to \textit{healthy} comments. Another limitation of the dataset is the low inter-reliability among the coders (Table~\ref{tab:dataset_preprocessed_krippendorff}). 
The low values of the metric reduce the confidence in the dataset, but also suggests that identification of different types of unhealthy conversations is challenging even for humans. Also, Krippendorff’s $\alpha$ was not reported for two out of eight classes (\textit{generalization unfair} and \textit{healthy}). In future work, we thus aim to expand our experiments on a more diverse dataset by replicating Prince et al.'s coding process using a variety of comments from different news websites.

One limitation of our machine learning analysis is that we trained a deep learning architecture (CNN-LSTM) using a relatively small dataset, mainly in terms of unhealthy categories of conversations. In future work, we plan to increase the number of unhealthy comments of our analysis by using data augmentation techniques such as Generative Adversarial Networks (GAN), which can synthetically generate new comments. NLP data augmentation techniques could also be used to increase the dataset without requiring significant human supervision, such as simulating keyboard distance error, substituting words according to their synonyms/antonyms, replacing words with their common spelling mistakes, etc.~\cite{csahin2019data}. This may help increase the performance of machine learning classifiers in general. 

Another limitation is that we included only four classifiers in our analysis, which may restrict the generalizability of our findings. Future work is therefore advised to include other models successfully used by previous studies in classifying conversations with unhealthy and toxic elements, such as RandomForest, Bi-LSTM networks, Stacked Bi-LSTM~\cite{Bagga}, capsule networks~\cite{Srivastava_Khurana_Tewari} and transformer models including BERT~\cite{Price_Gifford-Moore_Flemming_Musker_Roichman_Sylvain_Thain_Dixon_Sorensen_2020}. Future work is also advised to test feature selection methods, which usually help increase classification accuracy by focusing on the most discriminating features while discarding redundant and irrelevant information. 
Lastly, given the challenging nature of unhealthy comments classification tasks, future work should look at specialized corpora (e.g., \citealp{wang2019talkdown, oraby2017creating}) and machine learning models trained at differentiating particular types of conversations, for example, solely sarcastic comments from non-sarcastic ones (e.g., \citealp{pant2020sarcasm, hazarika-etal-2018-cascade}).

\section{Conclusion}\label{sec:conclusion}

This paper analyzed the granularities of subtle toxic online conversations. The study goal was to systematically investigate the general sentiment associated with healthy and unhealthy online comments, as well as the accuracy of machine learning classifiers in recognizing these comments. Towards this end, we leveraged a public dataset containing healthy and unhealthy comments labeled with seven forms of subtle toxicity. We were able to distinguish between these comments with a maximum micro F1-score, macro F1-score, and AUC-ROC of 88.76\%, 67.98\%, and 0.71, respectively, using a CNN-LSTM network with pre-trained word embeddings. Our conclusions are two-fold: (i) hostile comments are the most negative (i.e., less subtle toxic) and detectable form of unhealthy online conversation; (ii) most types of unhealthy comments are associated with a slight negative sentiment. 
Findings from this work have the potential to inform and advance future research and development of online moderation tools, which pave the way for safer online environments.

\newpage
\newpage


\bibliographystyle{acl_natbib}
\bibliography{sections/main.bib}

\begin{thebibliography}{55}
\expandafter\ifx\csname natexlab\endcsname\relax\def\natexlab#1{#1}\fi

\bibitem[{Abadi et~al.(2016)Abadi, Agarwal, Barham, Brevdo, Chen, Citro,
  Corrado, Davis, Dean, Devin, Ghemawat, Goodfellow, Harp, Irving, Isard, Jia,
  Jozefowicz, Kaiser, Kudlur, Levenberg, Mane, Monga, Moore, Murray, Olah,
  Schuster, Shlens, Steiner, Sutskever, Talwar, Tucker, Vanhoucke, Vasudevan,
  Viegas, Vinyals, Warden, Wattenberg, Wicke, Yu, and
  Zheng}]{tensorflow2015-whitepaper}
Martín Abadi, Ashish Agarwal, Paul Barham, Eugene Brevdo, Zhifeng Chen, Craig
  Citro, Greg~S. Corrado, Andy Davis, Jeffrey Dean, Matthieu Devin, Sanjay
  Ghemawat, Ian Goodfellow, Andrew Harp, Geoffrey Irving, Michael Isard,
  Yangqing Jia, Rafal Jozefowicz, Lukasz Kaiser, Manjunath Kudlur, Josh
  Levenberg, Dan Mane, Rajat Monga, Sherry Moore, Derek Murray, Chris Olah,
  Mike Schuster, Jonathon Shlens, Benoit Steiner, Ilya Sutskever, Kunal Talwar,
  Paul Tucker, Vincent Vanhoucke, Vijay Vasudevan, Fernanda Viegas, Oriol
  Vinyals, Pete Warden, Martin Wattenberg, Martin Wicke, Yuan Yu, and Xiaoqiang
  Zheng. 2016.
\newblock \href {https://www.tensorflow.org/} {{TensorFlow}: Large-scale
  machine learning on heterogeneous systems}.
\newblock \emph{arXiv preprint arXiv:1603.04467}.
\newblock Software available from tensorflow.org.

\bibitem[{van Aken et~al.(2018)van Aken, Risch, Krestel, and
  Löser}]{van_Aken_Risch_Krestel_Loser_2018}
Betty van Aken, Julian Risch, Ralf Krestel, and Alexander Löser. 2018.
\newblock \href {http://arxiv.org/abs/1809.07572} {Challenges for toxic comment
  classification: An in-depth error analysis}.
\newblock \emph{arXiv:1809.07572 [cs]}.
\newblock ArXiv: 1809.07572.

\bibitem[{Bagga(2020)}]{Bagga}
Sunyam Bagga. 2020.
\newblock Detecting abuse on the internet: It’s subtle.
\newblock \emph{McGill University}, page 105.

\bibitem[{Barbier et~al.(2016)Barbier, Dia, Macris, Krzakala, Lesieur, and
  Zdeborová}]{barbier_Dia_Macris_Krzakala_Lesieur_Zdeborova_2016}
Jean Barbier, Mohamad Dia, Nicolas Macris, Florent Krzakala, Thibault Lesieur,
  and Lenka Zdeborová. 2016.
\newblock \href
  {https://proceedings.neurips.cc/paper/2016/file/621bf66ddb7c962aa0d22ac97d69b793-Paper.pdf}
  {Mutual information for symmetric rank-one matrix estimation: A proof of the
  replica formula}.
\newblock In \emph{Advances in Neural Information Processing Systems},
  volume~29. Curran Associates, Inc.

\bibitem[{Blackwell et~al.(2017)Blackwell, Dimond, Schoenebeck, and
  Lampe}]{Blackwell_Dimond_Schoenebeck_Lampe_2017}
Lindsay Blackwell, Jill Dimond, Sarita Schoenebeck, and Cliff Lampe. 2017.
\newblock \href {https://doi.org/10.1145/3134659} {Classification and its
  consequences for online harassment: Design insights from heartmob}.
\newblock \emph{Proceedings of the ACM on Human-Computer Interaction},
  1(CSCW):24:1--24:19.

\bibitem[{Chen and Qian(2019)}]{Chen_Qian_2019}
Zhuang Chen and Tieyun Qian. 2019.
\newblock \href {https://doi.org/10.18653/v1/P19-1052} {Transfer capsule
  network for aspect level sentiment classification}.
\newblock In \emph{Proceedings of the 57\textsuperscript{th} Annual Meeting of
  the Association for Computational Linguistics}, pages 547--556. Association
  for Computational Linguistics.

\bibitem[{Chu et~al.(2016)Chu, Jue, and Wang}]{Chu2017CommentAC}
Theodora Chu, Kylie Jue, and Max Wang. 2016.
\newblock Comment abuse classification with deep learning.
\newblock \emph{Stanford University}.

\bibitem[{Dadu and Pant(2020)}]{pant2020sarcasm}
Tanvi Dadu and Kartikey Pant. 2020.
\newblock Sarcasm detection using context separators in online discourse.
\newblock In \emph{Proceedings of the Second Workshop on Figurative Language
  Processing}, pages 51--55.

\bibitem[{Davidson et~al.(2017)Davidson, Warmsley, Macy, and
  Weber}]{Davidson_Warmsley_Macy_Weber_2017}
Thomas Davidson, Dana Warmsley, Michael Macy, and Ingmar Weber. 2017.
\newblock \href {http://arxiv.org/abs/1703.04009} {Automated hate speech
  detection and the problem of offensive language}.
\newblock \emph{arXiv:1703.04009 [cs]}.
\newblock ArXiv: 1703.04009.

\bibitem[{Dinakar et~al.(2012)Dinakar, Jones, Havasi, Lieberman, and
  Picard}]{Dinakar_Jones_Havasi_Lieberman_Picard_2012}
Karthik Dinakar, Birago Jones, Catherine Havasi, Henry Lieberman, and Rosalind
  Picard. 2012.
\newblock \href {https://doi.org/10.1145/2362394.2362400} {Common sense
  reasoning for detection, prevention, and mitigation of cyberbullying}.
\newblock \emph{ACM Transactions on Interactive Intelligent Systems},
  2(3):1–30.

\bibitem[{Georgakopoulos et~al.(2018)Georgakopoulos, Tasoulis, Vrahatis, and
  Plagianakos}]{Georgakopoulos_Tasoulis_Vrahatis_Plagianakos_2018}
Spiros~V. Georgakopoulos, Sotiris~K. Tasoulis, Aristidis~G. Vrahatis, and
  Vassilis~P. Plagianakos. 2018.
\newblock \href {http://arxiv.org/abs/1802.09957} {Convolutional neural
  networks for toxic comment classification}.
\newblock \emph{arXiv:1802.09957 [cs]}.
\newblock ArXiv: 1802.09957.

\bibitem[{Glick and Fiske(2001)}]{glick2001ambivalent}
Peter Glick and Susan~T. Fiske. 2001.
\newblock \href {https://doi.org/10.1037/0003-066x.56.2.109} {An ambivalent
  alliance: Hostile and benevolent sexism as complementary justifications for
  gender inequality.}
\newblock \emph{American psychologist}, 56(2):109--118.

\bibitem[{Guberman and Hemphill(2017)}]{Guberman_Hemphill_2017}
Joshua Guberman and Libby Hemphill. 2017.
\newblock \href {https://doi.org/10.24251/hicss.2017.267} {Challenges in
  modifying existing scales for detecting harassment in individual tweets}.
\newblock In \emph{Proceedings of 50th Annual Hawaii International Conference
  on System Sciences ({HICSS})}.

\bibitem[{Han et~al.(2011)Han, Kamber, and Pei}]{han2012mining}
Jiawei Han, Micheline Kamber, and Jian Pei. 2011.
\newblock \href
  {http://www.amazon.de/Data-Mining-Concepts-Techniques-Management/dp/0123814790/ref=tmm_hrd_title_0?ie=UTF8&qid=1366039033&sr=1-1}
  {Data mining concepts and techniques, third edition}.
\newblock \emph{The Morgan Kaufmann Series in Data Management Systems},
  5(4):83--124.

\bibitem[{Hazarika et~al.(2018)Hazarika, Poria, Gorantla, Cambria, Zimmermann,
  and Mihalcea}]{hazarika-etal-2018-cascade}
Devamanyu Hazarika, Soujanya Poria, Sruthi Gorantla, Erik Cambria, Roger
  Zimmermann, and Rada Mihalcea. 2018.
\newblock \href {https://www.aclweb.org/anthology/C18-1156} {{CASCADE}:
  Contextual sarcasm detection in online discussion forums}.
\newblock In \emph{Proceedings of the 27th International Conference on
  Computational Linguistics}, pages 1837--1848, Santa Fe, New Mexico, USA.
  Association for Computational Linguistics.

\bibitem[{Hutto and Gilbert(2015)}]{inproceedings_vader}
C.~J. Hutto and Eric Gilbert. 2015.
\newblock Vader: A parsimonious rule-based model for sentiment analysis of
  social media text.
\newblock In \emph{Proceedings of the 8\textsuperscript{th} International
  Conference on Weblogs and Social Media, ICWSM 2014}.

\bibitem[{Jha and Mamidi(2017)}]{Jha_Mamidi_2017}
Akshita Jha and Radhika Mamidi. 2017.
\newblock \href {https://doi.org/10.18653/v1/W17-2902} {When does a compliment
  become sexist? analysis and classification of ambivalent sexism using twitter
  data}.
\newblock In \emph{Proceedings of the Second Workshop on NLP and Computational
  Social Science}, page 7–16. Association for Computational Linguistics.

\bibitem[{Jurgens et~al.(2019)Jurgens, Chandrasekharan, and
  Hemphill}]{Jurgens_Chandrasekharan_Hemphill_2019}
David Jurgens, Eshwar Chandrasekharan, and Libby Hemphill. 2019.
\newblock \href {http://arxiv.org/abs/1906.01738} {A just and comprehensive
  strategy for using nlp to address online abuse}.
\newblock \emph{arXiv:1906.01738 [cs]}.
\newblock ArXiv: 1906.01738.

\bibitem[{Ke et~al.(2017)Ke, Meng, Finley, Wang, Chen, Ma, Ye, and
  Liu}]{Ke_Meng_Finley_Wang_Chen_Ma_Ye_Liu_2017}
Guolin Ke, Qi~Meng, Thomas Finley, Taifeng Wang, Wei Chen, Weidong Ma, Qiwei
  Ye, and Tie-Yan Liu. 2017.
\newblock \href
  {https://proceedings.neurips.cc/paper/2017/file/6449f44a102fde848669bdd9eb6b76fa-Paper.pdf}
  {Lightgbm: A highly efficient gradient boosting decision tree}.
\newblock In \emph{Advances in Neural Information Processing Systems},
  volume~30. Curran Associates, Inc.

\bibitem[{Kolhatkar et~al.(2020)Kolhatkar, Wu, Cavasso, Francis, Shukla, and
  Taboada}]{Kolhatkar_Wu_Cavasso_Francis_Shukla_Taboada_2020}
Varada Kolhatkar, Hanhan Wu, Luca Cavasso, Emilie Francis, Kavan Shukla, and
  Maite Taboada. 2020.
\newblock \href {https://doi.org/10.1007/s41701-019-00065-w} {The sfu opinion
  and comments corpus: A corpus for the analysis of online news comments}.
\newblock \emph{Corpus Pragmatics}, 4(2):155–190.

\bibitem[{Le and Mikolov(2014)}]{Le_Mikolov_2014}
Quoc~V. Le and Tomas Mikolov. 2014.
\newblock \href {http://arxiv.org/abs/1405.4053} {Distributed representations
  of sentences and documents}.
\newblock \emph{arXiv:1405.4053 [cs]}.
\newblock ArXiv: 1405.4053.

\bibitem[{Liu et~al.(2018)Liu, Guberman, Hemphill, and
  Culotta}]{Liu_Guberman_Hemphill_Culotta_2018}
Ping Liu, Joshua Guberman, Libby Hemphill, and Aron Culotta. 2018.
\newblock \href {http://arxiv.org/abs/1804.06759} {Forecasting the presence and
  intensity of hostility on instagram using linguistic and social features}.
\newblock \emph{arXiv:1804.06759 [cs]}.
\newblock ArXiv: 1804.06759.

\bibitem[{Manning et~al.(2008)Manning, Raghavan, and Sch\"{u}tze}]{ir:2008}
Christopher~D. Manning, Prabhakar Raghavan, and Hinrich Sch\"{u}tze. 2008.
\newblock \emph{Introduction to Information Retrieval}.
\newblock Cambridge University Press, New York, NY, USA.

\bibitem[{Markey(2000)}]{Markey_2000}
P.~M. Markey. 2000.
\newblock \href {https://doi.org/https://doi.org/10.1016/S0747-5632(99)00056-4}
  {Bystander intervention in computer-mediated communication}.
\newblock \emph{Computers in Human Behavior}, 16(2):183–188.

\bibitem[{Mikolov et~al.(2013)Mikolov, Chen, Corrado, and
  Dean}]{Mikolov_Chen_Corrado_Dean_2013}
Tomas Mikolov, Kai Chen, Greg Corrado, and Jeffrey Dean. 2013.
\newblock \href {http://arxiv.org/abs/1301.3781} {Efficient estimation of word
  representations in vector space}.
\newblock \emph{arXiv:1301.3781 [cs]}.
\newblock ArXiv: 1301.3781.

\bibitem[{Nadal et~al.(2014)Nadal, Griffin, Wong, Hamit, and
  Rasmus}]{Nadal_Griffin_Wong_Hamit_Rasmus_2014}
Kevin~L. Nadal, Katie~E. Griffin, Yinglee Wong, Sahran Hamit, and Morgan
  Rasmus. 2014.
\newblock \href
  {https://doi.org/https://doi.org/10.1002/j.1556-6676.2014.00130.x} {The
  impact of racial microaggressions on mental health: Counseling implications
  for clients of color}.
\newblock \emph{Journal of Counseling \& Development}, 92(1):57–66.

\bibitem[{Oraby et~al.(2017)Oraby, Harrison, Reed, Hernandez, Riloff, and
  Walker}]{oraby2017creating}
Shereen Oraby, Vrindavan Harrison, Lena Reed, Ernesto Hernandez, Ellen Riloff,
  and Marilyn Walker. 2017.
\newblock \href {http://arxiv.org/abs/1709.05404} {Creating and characterizing
  a diverse corpus of sarcasm in dialogue}.
\newblock \emph{arXiv preprint arXiv:1709.05404}.

\bibitem[{Parekh and Patel(2017)}]{Parekh_Patel_2017}
Pooja Parekh and Hetal Patel. 2017.
\newblock Toxic comment tools: A case study.
\newblock \emph{International Journal of Advanced Research in Computer
  Science}, 8(5).

\bibitem[{Pedregosa et~al.(2011)Pedregosa, Varoquaux, Gramfort, Michel,
  Thirion, Grisel, Blondel, Prettenhofer, Weiss, Dubourg, Vanderplas, Passos,
  Cournapeau, Brucher, Perrot, and Duchesnay}]{scikit-learn}
F.~Pedregosa, G.~Varoquaux, A.~Gramfort, V.~Michel, B.~Thirion, O.~Grisel,
  M.~Blondel, P.~Prettenhofer, R.~Weiss, V.~Dubourg, J.~Vanderplas, A.~Passos,
  D.~Cournapeau, M.~Brucher, M.~Perrot, and E.~Duchesnay. 2011.
\newblock Scikit-learn: Machine learning in {P}ython.
\newblock \emph{Journal of Machine Learning Research}, 12:2825--2830.

\bibitem[{Pennebaker et~al.(2001)Pennebaker, Francis, and Booth}]{pennebaker01}
James~W. Pennebaker, Martha~E. Francis, and Roger~J. Booth. 2001.
\newblock \emph{Linguistic Inquiry and Word Count}.
\newblock Lawerence Erlbaum Associates, Mahwah, NJ.

\bibitem[{Pennington et~al.(2014{\natexlab{a}})Pennington, Socher, and
  Manning}]{Pennington_Socher_Manning_2014}
Jeffrey Pennington, Richard Socher, and Christopher Manning.
  2014{\natexlab{a}}.
\newblock \href {https://doi.org/10.3115/v1/D14-1162} {Glove: Global vectors
  for word representation}.
\newblock In \emph{Proceedings of the 2014 Conference on Empirical Methods in
  Natural Language Processing (EMNLP)}, page 1532–1543. Association for
  Computational Linguistics.

\bibitem[{Pennington et~al.(2014{\natexlab{b}})Pennington, Socher, and
  Manning}]{pennington2014glove}
Jeffrey Pennington, Richard Socher, and Christopher~D. Manning.
  2014{\natexlab{b}}.
\newblock \href {http://www.aclweb.org/anthology/D14-1162} {Glove: Global
  vectors for word representation}.
\newblock In \emph{Empirical Methods in Natural Language Processing (EMNLP)},
  pages 1532--1543.

\bibitem[{Poria et~al.(2016)Poria, Cambria, Hazarika, and
  Vij}]{poria2017deeper}
Soujanya Poria, Erik Cambria, Devamanyu Hazarika, and Prateek Vij. 2016.
\newblock \href {http://arxiv.org/abs/1610.08815} {A deeper look into sarcastic
  tweets using deep convolutional neural networks}.
\newblock \emph{arXiv preprint arXiv:1610.08815}.

\bibitem[{Price et~al.(2020)Price, Gifford-Moore, Flemming, Musker, Roichman,
  Sylvain, Thain, Dixon, and
  Sorensen}]{Price_Gifford-Moore_Flemming_Musker_Roichman_Sylvain_Thain_Dixon_Sorensen_2020}
Ilan Price, Jordan Gifford-Moore, Jory Flemming, Saul Musker, Maayan Roichman,
  Guillaume Sylvain, Nithum Thain, Lucas Dixon, and Jeffrey Sorensen. 2020.
\newblock \href {http://arxiv.org/abs/2010.07410} {Six attributes of unhealthy
  conversation}.
\newblock \emph{arXiv:2010.07410 [cs]}.
\newblock ArXiv: 2010.07410.

\bibitem[{Pérez-Almendros et~al.(2020)Pérez-Almendros, Espinosa-Anke, and
  Schockaert}]{perezalmendros2020dont}
Carla Pérez-Almendros, Luis Espinosa-Anke, and Steven Schockaert. 2020.
\newblock \href {http://arxiv.org/abs/2011.08320} {Don't patronize me! an
  annotated dataset with patronizing and condescending language towards
  vulnerable communities}.
\newblock \emph{arXiv preprint arXiv:2011.08320}.

\bibitem[{Qiufen(2014)}]{qiufen2014understanding}
Yu~Qiufen. 2014.
\newblock Understanding the impact of culture on interpretation: A relevance
  theoretic perspective.
\newblock \emph{Intercultural Communication Studies}, 23(3).

\bibitem[{Saeed et~al.(2018)Saeed, Shahzad, and
  Kamiran}]{Saeed_Shahzad_Kamiran_2018}
H.~H. Saeed, K.~Shahzad, and F.~Kamiran. 2018.
\newblock \href {https://doi.org/10.1109/ICDMW.2018.00193} {Overlapping toxic
  sentiment classification using deep neural architectures}.
\newblock In \emph{2018 IEEE International Conference on Data Mining Workshops
  (ICDMW)}, page 1361–1366.

\bibitem[{Safi~Samghabadi et~al.(2017)Safi~Samghabadi, Maharjan, Sprague,
  Diaz-Sprague, and
  Solorio}]{Safi_Samghabadi_Maharjan_Sprague_Diaz-Sprague_Solorio_2017}
Niloofar Safi~Samghabadi, Suraj Maharjan, Alan Sprague, Raquel Diaz-Sprague,
  and Thamar Solorio. 2017.
\newblock \href {https://doi.org/10.18653/v1/W17-3010} {Detecting nastiness in
  social media}.
\newblock In \emph{Proceedings of the First Workshop on Abusive Language
  Online}, page 63–72. Association for Computational Linguistics.

\bibitem[{{\c{S}}ahin and Steedman(2019)}]{csahin2019data}
G{\"o}zde~G{\"u}l {\c{S}}ahin and Mark Steedman. 2019.
\newblock Data augmentation via dependency tree morphing for low-resource
  languages.
\newblock \emph{arXiv preprint arXiv:1903.09460}.

\bibitem[{Saif et~al.(2018)Saif, Medvedev, Medvedev, and
  Atanasova}]{Saif_Medvedev_Medvedev_Atanasova_2018}
Mujahed~A. Saif, Alexander~N. Medvedev, Maxim~A. Medvedev, and Todorka
  Atanasova. 2018.
\newblock \href {https://doi.org/10.1063/1.5082126} {Classification of online
  toxic comments using the logistic regression and neural networks models}.
\newblock \emph{AIP Conference Proceedings}, 2048(1):060011.

\bibitem[{Salminen et~al.(2018)Salminen, Almerekhi, Milenkovic, Jung, An, Kwak,
  and Jansen}]{Salminen_Almerekhi_Milenkovic_Jung_An_Kwak_Jansen_2019}
Joni Salminen, Hind Almerekhi, Milica Milenkovic, Soon-Gyo Jung, Jisun An,
  Haewoon Kwak, and Jim Jansen. 2018.
\newblock Anatomy of online hate: Developing a taxonomy and machine learning
  models for identifying and classifying hate in online news media.
\newblock In \emph{Proceedings of the International AAAI Conference on Web and
  Social Media}, volume~12.

\bibitem[{Sechidis et~al.(2011)Sechidis, Tsoumakas, and
  Vlahavas}]{sechidis2011stratification}
Konstantinos Sechidis, Grigorios Tsoumakas, and Ioannis Vlahavas. 2011.
\newblock On the stratification of multi-label data.
\newblock \emph{Machine Learning and Knowledge Discovery in Databases}, pages
  145--158.

\bibitem[{Sennrich et~al.(2016)Sennrich, Haddow, and Birch}]{politeness}
Rico Sennrich, Barry Haddow, and Alexandra Birch. 2016.
\newblock \href {https://doi.org/10.18653/v1/N16-1005} {Controlling politeness
  in neural machine translation via side constraints}.
\newblock In \emph{Proceedings of the 2016 Conference of the North American
  Chapter of the Association for Computational Linguistics: Human Language
  Technologies}, pages 35--40.

\bibitem[{Srivastava et~al.(2018)Srivastava, Khurana, and
  Tewari}]{Srivastava_Khurana_Tewari}
Saurabh Srivastava, Prerna Khurana, and Vartika Tewari. 2018.
\newblock \href {https://www.aclweb.org/anthology/W18-4412} {Identifying
  aggression and toxicity in comments using capsule network}.
\newblock In \emph{Proceedings of the First Workshop on Trolling, Aggression
  and Cyberbullying ({TRAC}-2018)}, pages 98--105, Santa Fe, New Mexico, USA.
  Association for Computational Linguistics.

\bibitem[{Sue et~al.(2007)Sue, Capodilupo, Torino, Bucceri, Holder, Nadal, and
  Esquilin}]{sue2007racial}
Derald~Wing Sue, Christina~M. Capodilupo, Gina~C. Torino, Jennifer~M. Bucceri,
  Aisha M.~B. Holder, Kevin~L. Nadal, and Marta Esquilin. 2007.
\newblock \href {https://doi.org/10.1037/0003-066x.62.4.271} {Racial
  microaggressions in everyday life: Implications for clinical practice.}
\newblock \emph{American psychologist}, 62(4):271--286.

\bibitem[{{Szyma{\'n}ski} and {Kajdanowicz}(2017)}]{2017arXiv170201460S}
P.~{Szyma{\'n}ski} and T.~{Kajdanowicz}. 2017.
\newblock {A scikit-based Python environment for performing multi-label
  classification}.
\newblock \emph{ArXiv e-prints}.

\bibitem[{Szymański and Kajdanowicz(2017)}]{pmlr-v74-szymanski17a}
Piotr Szymański and Tomasz Kajdanowicz. 2017.
\newblock A network perspective on stratification of multi-label data.
\newblock In \emph{Proceedings of the First International Workshop on Learning
  with Imbalanced Domains:Theory and Applications}, volume~74 of
  \emph{Proceedings of Machine Learning Research}, pages 22--35, ECML-PKDD,
  Skopje, Macedonia. PMLR.

\bibitem[{Vapnik(1998)}]{Vapnik1998}
Vladimir~N. Vapnik. 1998.
\newblock \emph{Statistical Learning Theory}.
\newblock Wiley-Interscience.

\bibitem[{Wang and Potts(2019)}]{wang2019talkdown}
Zijian Wang and Christopher Potts. 2019.
\newblock \href {http://arxiv.org/abs/1909.11272} {Talkdown: A corpus for
  condescension detection in context}.
\newblock \emph{arXiv preprint arXiv:1909.11272}.

\bibitem[{Waseem et~al.(2017)Waseem, Davidson, Warmsley, and
  Weber}]{Waseem_Davidson_Warmsley_Weber_2017}
Zeerak Waseem, Thomas Davidson, Dana Warmsley, and Ingmar Weber. 2017.
\newblock \href {https://doi.org/10.18653/v1/W17-3012} {Understanding abuse: A
  typology of abusive language detection subtasks}.
\newblock In \emph{Proceedings of the First Workshop on Abusive Language
  Online}, page 78–84. Association for Computational Linguistics.

\bibitem[{Yin et~al.(2009)Yin, Xue, Hong, Davison, Edwards, and
  Edwards}]{Yin_Xue_Hong_Davison_Edwards_Edwards_2009}
Dawei Yin, Zhenzhen Xue, Liangjie Hong, Brian Davison, April Edwards, and Lynne
  Edwards. 2009.
\newblock Detection of harassment on web 2.0.
\newblock \emph{Proceedings of the Content Analysis in the {WEB}}, 2:1--7.

\bibitem[{Yu et~al.(2017)Yu, Wang, Lai, and Zhang}]{Yu_Wang_Lai_Zhang_2017}
Liang-Chih Yu, Jin Wang, K.~Robert Lai, and Xuejie Zhang. 2017.
\newblock \href {https://doi.org/10.18653/v1/D17-1056} {Refining word
  embeddings for sentiment analysis}.
\newblock In \emph{Proceedings of the 2017 Conference on Empirical Methods in
  Natural Language Processing}, page 534–539. Association for Computational
  Linguistics.

\bibitem[{Zhang et~al.(2017)Zhang, Si, and Hsieh}]{zhang2017gpuacceleration}
Huan Zhang, Si~Si, and Cho-Jui Hsieh. 2017.
\newblock \href {http://arxiv.org/abs/1706.08359} {Gpu-acceleration for
  large-scale tree boosting}.
\newblock \emph{arXiv preprint arXiv:1706.08359}.

\bibitem[{Zhang et~al.(2016)Zhang, Zhang, and Fu}]{Zhang2016TweetSD}
Meishan Zhang, Yue Zhang, and G.~Fu. 2016.
\newblock Tweet sarcasm detection using deep neural network.
\newblock In \emph{COLING}.

\bibitem[{Zheng(2015)}]{Zheng_2015}
A.~Zheng. 2015.
\newblock \href {https://books.google.com/books?id=OFhauwEACAAJ}
  {\emph{Evaluating Machine Learning Models: A Beginner’s Guide to Key
  Concepts and Pitfalls}}.
\newblock O’Reilly Media.

\end{thebibliography}

\end{document}